\documentclass[10pt]{article}
\usepackage{graphicx}
\usepackage{enumitem}
\usepackage{amsmath}
\usepackage{amssymb}
\usepackage{smartdiagram}
\usepackage{natbib}
\usepackage[linesnumbered,ruled,vlined]{algorithm2e}

\def\be{\begin{equation}}
\def\ee{\end{equation}}
\def\ff#1{{\mbox{\boldmath{$#1$}}}}
\def\lam{\lambda}
\def\x{\ff{x}}

\def\argmin{\textrm{argmin}}

\def\brak#1{\left\{  \begin{array}{lllllllll} #1 \end{array} \right. }

\begin{document}

\title{A Generalized Evolutionary Metaheuristic (GEM) Algorithm for Engineering Optimization}
\author{Xin-She Yang \\
School of Science and Technology,  Middlesex University London, \\
The Burroughs, London NW4 4BT, United Kingdom. }
%% \date{email: x.yang@mdx.ac.uk}
\date{}

\maketitle
% \pagestyle{headings}
% \pagestyle{plain}
% \pagenumbering{arabic}

\abstract{Many optimization problems in engineering and industrial design applications can be formulated as optimization problems with highly nonlinear objectives, subject to multiple complex constraints. Solving such optimization problems requires sophisticated algorithms and optimization techniques. A major trend in recent years is the use of nature-inspired metaheustic algorithms (NIMA). Despite the popularity of nature-inspired metaheuristic algorithms, there are still some challenging issues and open problems to be resolved. Two main issues related to current NIMAs are: there are over 540 algorithms in the literature, and there is no unified framework to understand the search mechanisms of different algorithms. Therefore, this paper attempts to analyse some similarities and differences among different algorithms and then presents a generalized evolutionary metaheuristic (GEM) in an attempt to unify some of the existing algorithms.
After a brief discussion of some insights into nature-inspired algorithms and some open problems, we propose a generalized evolutionary metaheuristic  algorithm to unify more than 20 different algorithms so as to understand their main steps and search mechanisms. We then test the unified GEM using 15 test benchmarks to validate its performance. Finally,  further research topics are briefly discussed. }


\bigskip
\noindent {\color{blue}{\bf Citation Detail: } \\[7pt]
Xin-She Yang, A Generalized Evolutionary Metaheuristic (GEM) Algorithm for Engineering Optimization, {\it Cogent Engineering}, vol. 11, no. 1, Article 2364041 (2024). \\{}
https://doi.org/10.1080/23311916.2024.2364041
}

\section{Introduction}

Many design problems in engineering and industry can be formulated as
optimization problems subject to multiple nonlinear constraints. To solve such optimization problems, sophisticated optimization algorithms and techniques are often used. Traditional algorithms such as Newton-Raphson method are efficient, but they use derivatives and calculations of these derivatives, especially the second derivatives in a high-dimensional space, can be costly. In addition, such derivative-based algorithms are usually local search and the final solutions may depend on the starting the point if optimization problems are highly nonlinear and multimodal~\citep{Boyd2004,Yang2020Nature}.
An alternative approach is to use derivative-free algorithms and many evolutionary algorithms, especially recent nature-inspired algorithms, do not use derivatives~\citep{Kennedy1995PSO,Storn1997DE,Pham2005,Yang2020Rev}.
These nature-inspired metaheuristic algorithms are flexible and easy to implement, and yet they are usually very effective in solving various optimization problems in practice.

Algorithms have been important through history~\citep{Beer2016,Chabert1999,Schri2005}. There are a vast spectrum of algorithms in the literature, ranging from fundamental algorithms to combinatorial optimization techniques~\citep{Chabert1999,Schri2005}. In some special classes of optimization problems, effective algorithms exist for
linear programming~\citep{Karmarkar1984} and quadratic programming~\citep{Zdenek2009} as well as convex optimization~\citep{Berts2003,Boyd2004}. However, for nonlinear optimization problems, techniques vary and often approximations, heuristic algorithms and metaheuristic algorithms are needed. Even so, optimal solutions cannot always be obtained  for nonlinear optimization.

Metaheuristic algorithms are approximation optimization techniques, and they use some form of heuristics with trial and error and some form of memory and solution selections~\citep{Glover1986,Glover1997}. Most metaheuristic algorithms are evolution-based and/or nature-inspired. Evolution-based algorithms such as genetic algorithm~\citep{Holland1975,Goldberg1989} are often called evolutionary algorithms. Algorithms such as particle swarm optimization (PSO)~\citep{Kennedy1995PSO}, bees algorithm~\citep{Pham2005,Pham2009} and firefly algorithm~\citep{Yang2009FA} are often called swarm intelligence based algorithms~\citep{Kennedy2001Book}.

However, terminologies in this area are not well defined and different researchers may use different terminologies to refer to the same things. In this paper, we use nature-inspired algorithms to mean all the metaheuristic algorithms that are inspired by some forms of evolutionary characteristics in nature, being biological, behaviour, social, physical and chemical
characteristics ~\citep{Yang2020Nature,YangHe2019}. In this broad sense, almost all algorithms can be called nature-inspired algorithms, including bees algorithms~\citep{Pham2014,Pham2015}, PSO~\citep{Kennedy2001Book}, ant colony optimization, bat algorithm, flower pollination algorithm, cuckoo search algorithm, genetic algorithm, and many others.

Nature-inspired algorithms have become popular in recent years, and it is estimated that there are several hundred algorithms and variants in the current literature~\citep{Yang2020Nature}, and the relevant literature is expanding with more algorithms emerging  every year. An exhaustive review of metaheuristic algorithms by Rajwar et al. \citep{Rajwar2023} indicated that there are over 540 metaheuristic algorithms with over 350 of such algorithms that were developed in the last 10 years. Many such new variants have been developed based on different characteristics/species from nature, social interactions and/or artificial systems, or based on the hybridization of different algorithms or algorithmic components, or based on different strategies of selecting candidate solutions and information sharing characteristics~\citep{Mohamed2020, Rajwar2023,Zelinka2015}.

From the application perspective, nature-inspired algorithms have been shown that they can solve a wide range of optimization problems~\citep{Abdel2019,Bekdas2018,Pham2015,Osaba2016BA}, from continuous optimization~\citep{Pham2014} and engineering
design optimization problems~\citep{Bekdas2018} to combinatorial optimization problems~\citep{Ouaarab2014,Osaba2016BA,Osaba2017FA}, multi-robots systems~\citep{Rango2018,Palmieri2018} and many other applications~\citep{Zelinka2015,Gavvala2019,Mohamed2020,Rajwar2023}.

Despite the wide applications of nature-inspired algorithms, theoretical analysis in contrast lacks behind. Though there are some rigorous analyses concerning genetic algorithm~\citep{Greenh2000}, PSO~\citep{Clerc2002} and the bat algorithm~\citep{Chen2018}, however, many new algorithms have not been analyzed in detail. Ideally, a systematical analysis and review should be carried out in the similar way to convex analysis~\citep{Berts2003} and
convex optimization~\citep{Boyd2004}. In addition, since there are so many different algorithms, it is difficult to figure out what search mechanisms
can be effective in determining the performance of a specific algorithm.
Furthermore, some of these 540 algorithms can look very similar, either in terms of their search mechanisms or updating equations, though they may look very different on the surface. This often can cause confusion and frustration to readers and researchers to see what happens in this research community.
In fact, there are many open problems and unresolved issues concerning
nature-inspired metaheuristic algorithms~\citep{Yang2018SwarmRev,YangHe2019,Yang2020Rev,Rajwar2023}.

Therefore, the purpose of this paper is two-fold: outlining some of the challenging issues and open problems, and then developing a generalized evolutionary metaheuristic (GEM)
to unify many existing algorithms. The rest of the paper is organized as follows. Section 2 first provides some insights into nature-inspired computing and then outlines some of the open problems concerning nature-inspired algorithms. Section 3 presents a unified framework of more than 20 different algorithms so as to look all the relevant algorithms in the same set of mathematical equations. Section 4 discusses 15 benchmark functions and case studies, whereas Section 5 carries out some numerical
experiments to test and validate the generalized algorithm.
Finally, Section 6 concludes with a brief discussion of future research topics.

\section{Insights and Open Problems}

\begin{figure}
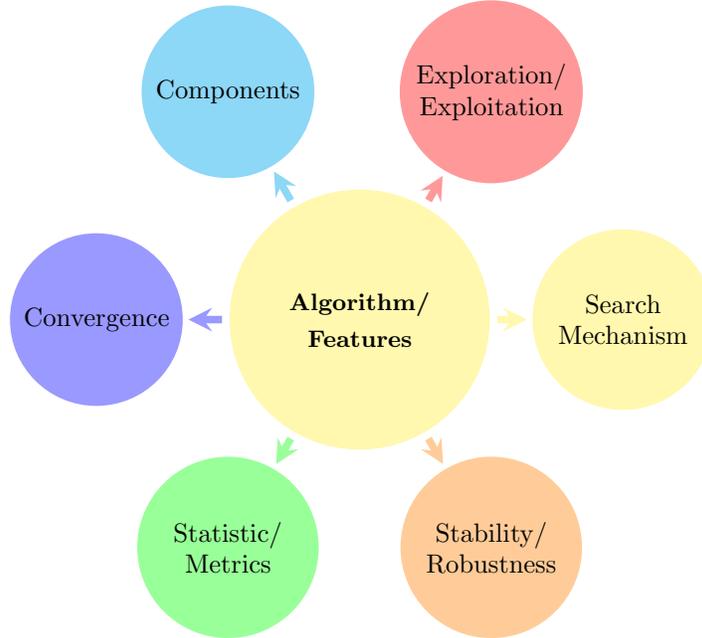

\centering
\smartdiagramset{planet text width=2cm, satellite text width=2cm,
planet size=1.7cm, planet color=yellow!40, satellite text opacity=0.9}
\smartdiagram[constellation diagram]{{\small \bf Algorithm/ Features},
Exploration/ Exploitation, Components,
Convergence,
Statistic/ Metrics, Stability/ Robustness,
Search Mechanism}

\caption{Analysis of algorithmic features such as components,
mechanisms and stability. \label{Alg-fig-100} }
\end{figure}

Though not much systematical analysis of nature-inspired algorithms exist, various studies from different perspectives have been carried out and different degrees of insights have been gained~\citep{Eiben2011,Yang2020Rev,Ser2019}. For any given nature-inspired algorithm, we can analyze its algorithmic components, and their role
in exploration and exploitation, and we also study its search mechanism so as to understand ways that local search and global search moves are carried out. Many studies in the literature have provided numerical convergence curves during iterations when solving different function optimization and sometimes real-world case studies, and such convergence curves are often presented with various statistical quantities according to a specific set of performance metrics such as accuracy of the solution and successful rate as well as number of iterations. In addition, stability and robustness are also studies for some algorithms~\citep{Clerc2002}. Such analyses, though very important, are largely qualitative studies of algorithmic features, as summarized
in Fig.~\ref{Alg-fig-100}.

\begin{figure}
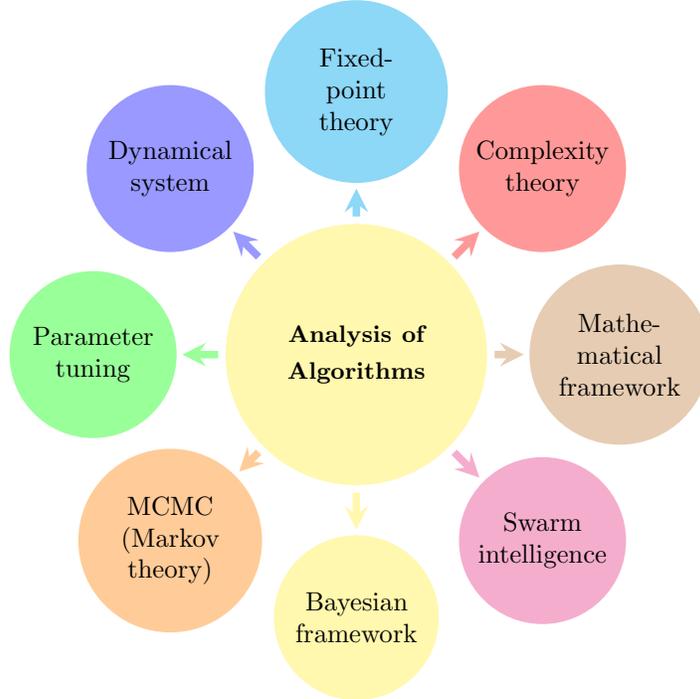

\centering
\smartdiagramset{planet text width=2cm, planet size=1.7cm,
planet color=yellow!40, satellite text width=1.77cm}
\smartdiagram[constellation diagram]{
{\small \bf Analysis of Algorithms},
Complexity theory, Fixed-point theory,
Dynamical system, Parameter tuning,
MCMC (Markov theory),
Bayesian framework, Swarm intelligence,
Mathematical framework}
\caption{Different perspectives for quantitative analysis of algorithms.
\label{Alg-fig-200} }
\end{figure}

The analysis of algorithms can be carried out more rigorously from a quantitative perspective, as shown in Fig.~\ref{Alg-fig-200}.  For a given algorithm, it is possible to analyze the iterative behaviour of the algorithm using fixed-point theory. However, the assumptions required for such theories  may not be realistic or relevant to the actual algorithm under consideration. Thus, it is not always possible
to carry out such analysis. One good way is to use complexity theory to analyze an algorithm to see its time complexity. Interestingly, most nature-inspired algorithms have the complexity of $O(n t)$ where $n$ is the
typically population size and $t$ is the number of iterations. It is still a mystery that how such low complexity algorithms can solve highly complex
nonlinear optimization problems that have been shown in various applications.

From the dynamical system point of view, an algorithm is a system of
updating equations, which can be formulated as a discrete dynamical system.
The eigenvalues of the main matrix of such a system determine the
main characteristics of the algorithm. It can be expected that these eigenvalues can depend on the parameter values of the algorithm, and thus
parameter settings can be important. In fact, the
analyses on PSO~\citep{Clerc2002} and the bat algorithm~\citep{Chen2018} show that parameter values are important. If the parameter values are in the wrong ranges, the algorithm may become unstable and become less effective.
This also indicates the important of parameter tuning in nature-inspired algorithms~\citep{Eiben2011,Joy2023Rev}. However, parameter tuning itself is
a challenging task because its aim is to find the optimal parameter setting for an optimization algorithm for a given set of optimization problems.

From the probability point of view, an algorithm can be considered as a
set of interacting Markov chains, thus it is possible to do some approximation analysis in terms of convergence using Markov chain Monte Carlo (MCMC) theory~\citep{Chen2018}. However, the conditions required for MCMC can be stringent, and thus not all algorithms can be analyzed in this way.
From the perspective of the analysis of variance, it is possible to
see how the variances may change with iteration to gain some
useful understanding~\citep{Zaharie2009}.

An alternative approach is to use Bayesian statistical framework to gain some insights into the algorithm under analysis. Loosely speaking, the initialization
of the population in an algorithm with a given probability distribution forms the prior of the algorithm. When the algorithm evolves and the solutions also evolve, leading to a posterior distribution of solutions and parameters. Thus, the evolution of algorithm can be understood from this perspective.
However, since Bayesian framework often requires a lot of extensive integral
evaluations, it is not straightforward to gain rigorous results in general.

A more ambition approach is to build a mathematical framework so as to analyze algorithms in a unified way, though such a framework does not exist in the current literature. Ideally, a theoretical framework should provide
enough insights into the rise of swarm intelligence, which is still an open
problem~\citep{Yang2020Rev,YangHe2019}.

As we have seen, algorithms can potentially be analyzed from different perspectives and there are many issues that need further research in the
general area of swarm intelligence and nature-inspired computation. We can highlight a few important open problems.

\begin{enumerate}
\item \emph{Theoretical framework}. Though there are some good theoretical analyses of a few algorithms such as genetic algorithm~\citep{Greenh2000}, PSO~\citep{Clerc2002} and the bat algorithm~\citep{Chen2018}, there is no unified theoretical framework that can be used to analyze all algorithms or at at least a major subset of nature-inspired algorithms. There is a strong need to build a mathematical framework so that the convergence and rate of convergence of any algorithm can be analyzed with rigorous and quantitative results.

    In addition, stability of an algorithm and its robustness should also be analyzed using the same mathematical framework, based on theories  such as dynamical systems and perturbation as well as probability. The insights gained from such a theoretical framework should provide enough guidance for tuning and setting parameter values for a given algorithm. However, how to construct this theoretical framework is still an open problem. It may be that a multidisciplinary approach is needed to ensure to look at algorithms from different perspectives.

\item \emph{Parameter tuning}. The setting of parameters in an algorithm can influence the performance of an algorithm, though the extent of
    such influence may largely depend on the algorithm itself and potentially on the problem to be solved~\citep{Joy2023Rev}.
    There are different methods for parameter tuning, but it is not clear
    which method(s) should be used for a given algorithm. In addition, different tuning methods may produce different results for parameter
    settings for different problems, which may leads to the question if a truly optimal parameter setting exists. It seems that there are
    different optimality conditions concerning parameter setting~\citep{Yang2013STA,Joy2023Rev}, and parameter settings may be
    both algorithm-dependent and problem-dependent, depending
    on the performance metric used for tuning. Many of these questions remain unresolved.

\item \emph{Benchmarking}. All new algorithms should be tested and validated using a diverse of set of benchmarks and case studies.
    In the current literature, one of the main issues is that most
    tests use smooth functions as benchmarks, and it seems that these
    functions have nothing to do with real-world applications. Thus, it
    is not clear how such tests can actually validate the algorithm to
    gain any insight into the potential performance of the algorithm
    to solve much more complicated real-world applications. There is
    a strong need to systematically investigate the role of benchmarking
    and what types of benchmarks and case studies should be used.

\item \emph{Performance metrics}. It can be expected that the performance of an algorithm depends on the performance metrics used to measure the performance. In the current literature, performance measures are mostly accuracy compared to the true objective values, success rate of multiple functions, number of iterations as computational efforts, computational time, and the combination of these measures. Whether these measures are fair or sufficient is still unexplained. In addition, these performance metrics tend to lead to the ranking of algorithms used and the benchmark functions used. Again, this may not be consistent with the no-free-lunch theorems~\citep{Wolpert1997}. It is not clear if other performance measures should be designed and used, and what theory should be based on to design performance metrics. All these are still
    open questions.

\item \emph{Search mechanism}. In many nature-inspired metaheuristic algorithms, certain forms of randomization and probability distributions are used to generate solutions with exploration abilities. One of the main tasks
    is to balance exploration and exploitation
    or diversification and intensification using different search
    moves or search mechanisms. However, how to balance these two components is still an open problem. In addition, exploration
    is usually by randomness, random walks and perturbations,
    whereas exploitation is usually by
    using derivative information and memory. It is not clear what
    search moves can be used to achieve both exploration and exploitation
    effectively.

\item \emph{Scalability}. Most studies of metaheuristic algorithms in the literature are concerned with problems with a few parameters or a few dozen parameters. These problems, though very complicated and useful,
    are small-scale problems. It is not clear if the algorithms used and the implementation realized can directly be applied to large-scale
    problems in practice. Simple scale up by using high-performance computing or cloud computing facilities may not be enough.
    How to scale up to solve real-world, large-scale problems is
    still a challenging issue. In fact, more efficient algorithms
    are always desirable in this context.

\item \emph{Rise of Swarm Intelligence}. Various discussions about
swarm intelligence have attracted attention in the current literature. It is not clear what swarm intelligence exactly means and what conditions are necessary to achieve such collective intelligence. There is a strong need to understand swarm intelligence theoretically and practically so as to gain insights into the rise of swarm intelligence. With newly gained
 insights, we may be able to design better and more effective algorithms.

\end{enumerate}

In addition, from both theoretical perspective and practical point of view,
the no-free lunch theorems~\citep{Wolpert1997} had some significant impact
on the understanding of algorithm behaviour. Studies also indicate that
free lunches may exist for co-evolution~\citep{Wolpert2005}, continuous problems~\citep{Auger2010FL} and multi-objective optimization~\citep{Corne2003FL,Zitzler2003} under certain conditions. The main question is how to use such possibilities to build more effective algorithms.

\section{A Generalized Evolutionary Metaheuristic (GEM) for Optimization}

From the numerical algorithm analysis point of view~\citep{Boyd2004,Yang2020Nature}, an algorithm in essence is a procedure to modify the current solution $\x^t$ so as to produce a
potentially better solution $\x^{t+1}$. The well-known Newton's method at iteration $t$ can be written as
\be \x^{t+1}=\x^t-\frac{\nabla f(\x^t)}{\nabla^2 f(\x^t)}, \ee
where $\nabla f$ is the gradient vector of the objective function at $\x^t$
and $\nabla^2 f$ is the Hessian matrix. Loosely speaking,
all iterative algorithms can schematically be written as
\be \x^{t+1} =\x^t+ \ff{S} (\x^t, \x^*, p_1, ..., p_k), \ee
where $\ff{S}$ is a step size vector, which can in general depend on
the current solution $\x^t$, the current best solution $\x^*$, and a set of parameters ($p_1, ..., p_k$). Different algorithms may differ only in the
ways of generating such steps.

In addition to the open problems highlighted above, the other
purpose of this paper is to provide a unified algorithm framework, based on
more than 20 existing nature-inspired algorithms. This unification
may enable us to understand the links and difference among different algorithms. It also allows us to build a single generic algorithm that can
potentially use the advantages of all its component algorithms,
leading to an effective algorithm.

As we have mentioned earlier, there are approximately 540 metaheuristic algorithms and variants in the literature \citep{Rajwar2023}, it is impossible to test all algorithms and try to provide a unified approach.
This paper is the first attempt to unify multiple algorithms in the same generalized evolutionary perspective. Thus, we have to select over twenty algorithms to see if we can achieve our aim. Obviously, there are two
challenging issues: how many algorithms we should use and which algorithms we should choose. For the former question, we think it makes sense that we should choose at least ten different algorithms or more so that we can get a reasonable picture of the unification capabilities. In fact, we have chosen 22 algorithms for this framework. As for the latter question which algorithms to use, it is almost impossible to decide what algorithms to use from 540 algorithms. In the end, the algorithms we have chosen here have different search features or characteristics in terms of their exploration and exploitation capabilities. In addition, in the case of similar exploration and exploitation capabilities, we tend to select the slightly well-established algorithms that appeared earlier in the literature because later/new algorithms may have been based on such mechanisms.

This unified algorithm is called Generalized Evolutionary
Metaheuristic (GEM) with multiple parameters and components
to unify more than 20 algorithms.

A solution vector $\x$ in the $D$-dimensional space is denoted by
\be \x=(x_1, x_2, ..., x_D). \ee
For a set of $n$ solution vectors $\x_i$ where $i=1,2,...,n$, these
vectors evolve with the iteration $t$ (the pseudo-time) and they are
denoted by $\x_i^t$. Among these $n$ solutions, there is one solution
$\ff{g}_*$ that gives the best objective value (i.e., highest for maximization or lowest for minimization). For minimization,
\be \ff{g}_*=\argmin \{ f(\x_i^t), f(\x_i^*), f(\bar \x_{g}) \}, \quad
\textrm{(for minimization),} \label{alg-eq-100} \ee
where the argmin is to find the corresponding solution vector with the
best objective value. Here, $\bar \x_g$ is the average of the top
$m$ best solutions among $n$ solutions ($m \le n$). That is
\be \bar \x_g=\frac{1}{m} \sum_{j=1}^m x_j. \ee
In case of $m=n$, this becomes the centre of the gravity of the
whole population. Thus, we can refer to this step as the {\it centrality
move}.

The main updating
equations for our proposed unified algorithm are guided randomization and position update. In the \emph{guided randomization search} step,
it consists of
\be \ff{v}_i^{t+1}=\underbrace{p \ff{v}_i^t}_{\textrm{inertia term}} + \underbrace{q \epsilon_1 (\ff{g}_*-\x_i^t)}_{\textrm{motion towards current best}}
+ \underbrace{r \epsilon_2 (\x_i^*-\x_i^t)}_{\textrm{motion to individual best}}. \label{alg-eq-200} \ee
In the \emph{position update} step, it means \be
\x_{i, new}^{t+1}= a \x_i^t + \underbrace{(1-a) \bar{\x}_g}_{\textrm{Centrality}} +\underbrace{ b (\x_j^t-\x_i^t)}_{\textrm{similarity convergence}} + \underbrace{c \ff{v}_i^{t+1}}_{\textrm{kinetic motion}}
+\underbrace{\Theta h(\x_i^t) \zeta_i}_{\textrm{perturbation}}, \label{alg-eq-300} \ee
where $a, b, c, \Theta$, $p$, $q$, and $r$ are parameters, and $\x_i^*$ is the individual best solution for agent or particle $i$. Here, $h(\x_i^t)$ is a function of current solution, and in most case we can set $h(\x_i^t)=1$,
which can be a constant.  In addition, $\zeta_i$ is a vector of random numbers, typically drawn from a standard normal distribution. That is
\be \zeta_i \sim {\cal N}(0, 1). \ee
It is worth pointing out the effect of each term can be different, and thus we can name each term, based on their effect and meaning. The inertia term
is $p \ff{v}_i^t$, whereas the $q \epsilon_1 (\ff{g}_*-\x_i^t)$ simulates
the motion or move towards the current best solution. The term $r \epsilon_2 (\x_i^*-\x_i^t)$ means to move towards the individual best solution.
The centrality term has a weighting coefficient $(1-a)$, which tries to balance between the importance of the current solution $\x_i^t$ and the importance of the centre of gravity of the swarm $\bar{\x}_g$.
The main term $b (\x_j^t-\x_i^t)$ shows similar solutions should converge
with subtle or minor changes. $c \ff{v}_i^t$ is the kinetic motion of each solution vector. The perturbation term $\Theta h(\x_i^t) \zeta_i$ is controlled by the strength parameter $\Theta$ where $h(\x_i)$ can be used to
simulate certain specialized moves for each solution agent. If there is no specialization needed, $h(\x_i)=1$ can be used.

The {\it selection and acceptance criterion} for minimization is
\be \x_i^{t+1} =\brak{ \x_{i, new}^t & \textrm{if } \; f(\x_{i, new}^{t+1}) \le f(\x_i^t), \\ \\ \x_i^t & \textrm{otherwise.} } \label{alg-eq-400} \ee

These four main steps/equations are summarized in the pseudocode, shown
in Algorithm~\ref{Alg-ps-100} where the initialization of the population follows the standard Monte Carlo randomization
\be \x_i=Lb + rand(1,D) (Ub-Lb), \label{Init-Eq-100} \ee
where Lb and Ub are, respectively, the lower bound and upper bound of the
feasible decision variables. In addition, rand(1,D) is a vector of $D$ random numbers drawn from a uniform distribution.

Special cases can correspond to more than 20 different algorithms. It is worth pointing out that in many case there are more than one way to represent the same algorithm by setting different values of parameters. The following representations are one of the possible ways for representing these algorithms:
\begin{enumerate}
\item Differential evolution (DE)~\citep{Storn1997DE,Price2005DE}: $a=1$, $b=F$, $c=0$ and $\Theta=0$.

\item Particle swarm optimization (PSO)~\citep{Kennedy1995PSO,Kennedy2001Book}: $a=1$, $b=0$, $p=1$, $c=1$, and $\Theta=0$. In addition, $q=\alpha$ and $r=\beta$.

\item Firefly algorithm (FA)~\citep{Yang2009FA,Yang2013CSFA}: $a=1$, $b=\beta \exp(-\gamma r_{ij}^2)$,
$c=0$ and $\Theta=0.97^t$.

\item Simulated annealing (SA)~\citep{Kirk1983SA}: $a=1$, $b=c=0$ and $\Theta=1$.

\item Artificial bee algorithm (ABC)~\citep{Karaboga2008}: $a=1$, $b=\Phi \in [-1,1]$, $c=0$ and $\Theta=0$.

\item Artificial cooperative search (ACS)~\citep{Civic2013}: $a=1$, $b=R$. $c=0$, $\Theta=0$ and its two predator keys are drawn from its $\alpha$ and $\beta$.

\item Charged system search (CSS)~\citep{Kaveh2010ChSS}: $a=1$, $b=A(R)$ where $R$ is the normalized distance with the maximum at $R=1$. In addition, $c=0$
    and $\Theta=0$.

\item Cuckoo search (CS)~\citep{YangCS2010}: $a=1$, $c=0$, and $b=\alpha s H(p_a-\epsilon)$ where $\alpha$ is a scaling parameter, $H$ is a Heaviside function with a switch probability $p_a$ and a uniformly distributed random number $\epsilon$. The step size $s$ is drawn from a L\'evy distribution. The other branch with the switch probability $p_a$ is $a=1$, $b=c=0$, $\Theta=1$, but $\zeta$ is drawn form the
    L\'evy distribution
    \[ L(s, \beta) \sim \frac{\beta \Gamma(\beta) \sin (\pi \beta/2)}{\pi} \frac{1}{s^{1+\beta}}, \quad \beta=1.5, \]
    where $\Gamma$ is the standard Gamma function.

\item Gravitational search algorithm (GSA)~\citep{Rashedi2009GSA}: $a=1$, $p=$rand$\in [0,1]$,
$q=r=0$, $c=1$ and $\Theta=0$. In addition, $b=$rand$*G(t)$ where
$G(t)=G_0 \exp(-\alpha t/T)$.

\item Gradient evolution algorithm (GEA)~\citep{Kuo2015}: $a=1$, $c=0$ and
  $\Theta=0$, but \[
  b=r_g \Delta x_{ij}^t/{2(x_{ij}^w-x^t_{ij}+x^B_{ij})}. \] In addition, the parameter is set to $r_a=0$ in GEA.

\item Harris hawk optimizer (HHO)~\citep{Heidari2019}: $a=1$, $b=0$, $c=1$,
$\Theta=0$, $p=0$, $r=0$, and $q=-E$ where $E$ varies with iteration $t$.

\item Henry gas solubility optimization (HGSO)~\citep{Hashim2019}: $a=1$, $b=r$, $c=1$ and $\Theta=0$. In addition, $p=1$ and $r=0$, but $q$ is related to the solubility affected by the Henry constant and parochial pressure. Here, the Henry's constant changes with iteration $t$ by  $H^{t+1}_j=H^t_j \exp[-C_j (1/T(t)-1/T_0)]$ where $T_0=298.15$ K,  $C_j$ is a given constant and $T(t)=e^{-t/t_{\max}}$.

\item Harmony search (HS)~\citep{Geem2001}: $a=0$, $b=c=0$, $\Theta=1$, $h(\x)=\x_i^t$ for pitch adjustment, but for harmony selection $b=1$ and $\Theta=0$ can be used.

\item Ant lion optimizer (ALO)~\citep{Mirjalili2015}: $a=1$, $b=0$, $c=1$,
$\Theta=d_i-c_i$ where $d_i$ is the scaled random walk length and $c_i$
is its corresponding minimum.  In addition, $p=0$, $q=0$, $r=1$ with some variation of $\ff{g}_*$ is just the average of two selected elite ant-lion solutions.

\item Whale optimization algorithm (WOA)~\citep{Mirja2016Whale}: $a=1$, $c=0$, $p=q=r=0$, $b=-A$ where $A=2d r-a$ with $r$ be drawn randomly from $[0,1]$
    and $d$ be linearly decreasing from $2$ to $0$. In their spiral branch, $\Theta=1$ with an equivalent $b=e^{R} \cos(2 \pi R)$ where $R$ is uniformly distributed in [-1, 1].

\item Lion optimization algorithm (LOA)~\citep{Yaz2016}: $a=1$, $b=0$,
$c=1$, $\Theta=0$, $p=0$, $q=-PI$, and $r=0$. In the other moving branch in their LOA, $\Theta$ is proportional to $D R$ where
$D$ is their scaled distance and $R$ can be drawn either from [0, 1] or [-1, 1].

\item Mayfly optimization algorithm (MOA)~\citep{Zerv2020}:
$a=1$, $b=0$, $c=1$, $\Theta=0$, $p=g$, $q=a_1 \exp(-\beta r_p^2)$,
and $r=a_2 \exp(-\beta r_g^2)$.

\item Big bang-big crunch (BBBC)~\citep{Erol2006}: the modification of solutions is mainly around the centre of gravity $\x_g$ of the population with $a=0$.  $b=c=0$, and $\Theta=Ub/t_{\max}$.

\item Social spider algorithm (SSA)~\citep{James2015}: $a=1$, $b=w e^{d_{ij}^2}$ where $w$ is a weight coefficient and $d_{ij}$ is the distance between two spiders. In addition, $c=0$, $\Theta=1$, $\zeta=rand-1/2$.

\item Moth search algorithm (MSA)~\citep{Wang2018}: $a=\lambda$, $b=0$,
$c=1$, $p=0$, $q=\phi$, $r=0$, and $\Theta=0$ in one equation. The other
equation corresponds to $a=1$, $b=c=0$, $\Theta=S_{\max}/t^2$ and $\zeta=L(s)$ drawn from a L\'evy distribution.

\item  Multi-verse optimizer (MVO)~\citep{Mirja2016}: $a=1$, $b=c=0$, but $\Theta=1-(t/t_{\max})^{1/p}$ where $t_{\max}$ is the maximum number of iterations. Its randomized perturbation is given by $\zeta=\pm [Lb+rand (Ub-Lb)]$.

\item Water cycle algorithm (WCA)~\citep{Eskandar2012}: $a=1$, $c=\Theta=0$, $b=C \, rand$ where $C \in [1, 2]$ and $rand$ is a uniformly distributed random number in [0, 1] for both water drops in river and stream in the WCA. For the additional search for the new stream step, $a=1$, $b=c=0$, but $\Theta=\sqrt{\mu}$ as its standard deviation and $\zeta$ is drawn from a Gaussian distribution with a unity mean.

\end{enumerate}

\setlength{\algomargin}{15pt}
\begin{algorithm}
\caption{GEM: A generalized evolutionary metaheuristic for optimization.
\label{Alg-ps-100} }
\SetAlgoLined
\KwData{Define optimization objective ($f(\x)$) with proper constraint handling}
\KwResult{Best or optimal solution ($f_{\min}$) with archived solution history}

Initialize parameter values ($n$, $a$, $b$, $c$, $\Theta$, $p$, $q$, $r$)\;
Generate an initial population of $n$ solutions using Eq.~\eqref{Init-Eq-100}\;
Find the initial best solution $\ff{g}_*$ and $f_{\min}$ among the initial population\;

\While {($t<$MaxGeneration)}
{  Generate random numbers ($\epsilon_1, \epsilon_2, \zeta_i$)\;
   \For {$i=1:n$ (all solutions)}{
   Modify solutions by Eq.~(\ref{alg-eq-200})\;
   Update solution vectors by Eq.~(\ref{alg-eq-300})\;
   Accept a solution by checking criterion Eq.~(\ref{alg-eq-400})\;
  }

  Rank all the solutions to find the current best $f_{\min}$\;
  Update the best solution so far $\ff{g}_*$ and $\bar \x_g$ by Eq.~(\ref{alg-eq-100})\;
  Save and archive the current population for next generation\;
  Update the iteration counter $t \leftarrow t+1$\;
}

Post-process the solutions for constraint verification and visualization\;
\end{algorithm}

As pointed out earlier, there are more than one ways of representing an algorithm under consideration using the unified framework, and the minor
details and un-important components of some algorithms may be ignored.
In essence, the unified framework intends to extract the key components
of multiple algorithms so that we can figure out what main search
mechanisms or moves are important in metaheuristic algorithms.
Therefore, it can be expected that many other algorithms may
be considered as special cases of the GEM framework if the
right combinations of parameter values are used.

\section{Benchmarks and Parameter Settings}

To test the unified GEM algorithm, we have selected 15 benchmark functions
and case studies. There are many different test function benchmarks,
including multivariate functions~\citep{JamilYang2013}, CEC2005 test suite~\citep{Suganthan2005}, unconstrained benchmark function repository~\citep{Al-Roomi2015} and engineering optimization problems~\citep{Cag2008,Coello2000Con}. The intention is to select a subset
of optimization problems with a diverse range of properties such as
modality, convexity, nonlinear constraints, separability, and landscape variations. The case studies also include a mixed-integer programming
pressure vessel design problem, and a parameter estimation based on
data for a vibrations system governed by an ordinary differential equation (ODE).

\subsection{Test Functions}

The ten test functions are outlined as follows.
\begin{enumerate}
\item Sphere function
\be f_1(\x)=\sum_{i=1}^D x_i^2, \quad -10 \le x_i \le +10, \ee
its global optimality $f_{\min}=0$ occurs at $\x_*=(0, 0, ...,0)$.

\item Rosenbrock function~\citep{Rosenbrock1960}
\be f_2(\x)=(x_1-1)^2+100 \sum_{i=1}^{D-1} (x_{i+1}-x_i^2)^2,
\quad x_i \in [-10, 10], \ee
whose global minimum $f_{\min}=0$ is located at $\x_*=(1,1,...,1)$.

\item Ackley function
\be f_3(\x)=-20 \exp\left[-0.2 \sqrt{\frac{1}{D}
\sum_{i=1}^D x_i^2} \, \right]
-\exp\left[\frac{1}{D} \sum_{i=1}^D \cos(2 \pi x_i) \right]
+20 +e, \ee
with
\be x_i \in [-32.768, 32.768], \ee
whose global minimum $f_{\min}=0$ occurs at $\x_*=(0,0, ..., 0)$.

\item Dixon-Price function
\be f_4(\x)=(x_1-1)^2 + \sum_{i=2}^D i (2 x_i^2-x_{i-1})^2, \quad
x_i \in [-10, 10], \ee
whose global minimum $f_{\min}=0$ at $x_i=2^{-(2^i-2)/2^i}$ for $i=1, 2, ..., D$.

\item Schwefel function~\citep{Schwefel1995}
\be f_5(\x)=-x_1 x_2 (72-2x_1-2x_2), \quad x_i \in [0, 500],
\ee
its global minimum $f_{\min}=-3456$ occurs at $\x_*=(12, 12)$.

\item Booth function
\be f_6(\x)=(x_1+2x_2-7)^2 + (2x_1+x_2-5)^2, \quad
x_i \in [-10, 10], \ee
whose global minimum $f_{\min}=0$ occurs at $\x_*=(1, 3)$.

\item Holder table function
\be f_7(\x)=-\left|\sin(x_1) \cos(x_2) e^{\left|1-\frac{\sqrt{x_1^2+x_2^2}}{\pi} \right|} \right|,
\quad x_i \in [-10, 10],
 \ee
whose four global minima $f_{\min}=-19.2085$ occur
at $\x_*=(\pm 8.05502, \pm 9.66459)$.

\item Beale function
\be f_8(\x)=(1.5-x_1+x_1 x_2)^2+(2.25-x_1 +x_1 x_2^2)^2
+(2.625-x_1 +x_1 x_2^3)^2, \ee
where \be x_i \in [-4.5, +4.5]. \ee
The global minimum $f_{\min}=0$ occurs at $\x_*=(3, 0.5)$.

\item Trid function
\be f_9(\x)=\sum_{i=1}^D (x_i-1)^2 - \sum_{i=2}^D x_i x_{i-1},
\quad x_i \in [-d^2, d^2], \ee
whose global minimum $f_{\min}=-D(D+4)(D-1)/6$ occurs at
$x_i=i (D+1-i)$ for $i=1, 2, ..., D$.

\item Rastrigin function
\be f_{10}(\x)=10 D + \sum_{i=1}^D [x_i^2-10 \cos(2 \pi x_i)], \quad
x_i \in [-5.12, 5.12], \ee
whose global minimum $f_{\min}=0$ occurs at
$\x_*=(0, 0, ...,0)$.

\end{enumerate}

\subsection{Case Studies}

The five design case studies or benchmarks are described as follows.

\begin{enumerate}[resume]

\item Spring design. The three design variables are: the wire diameter $w$ (or $x_1$), mean coil diameter $D$ (or $x_2$) and
the number $N$ (or $x_3$) of active coils.

\be \min \; f(\x)=(2+x_3) x_1^2 x_2, \ee
subject to
\begin{align}
& g_1(\x)=1-\frac{x_2^3 x_3}{71785 x_1^4} \le 0, \\
& g_2(\x)=\frac{4 x_2^2-x_1 x_2}{12566 (x_2 x_1^3-x_1^4)}+\frac{1}{5108 x_1^2}-1 \le 0, \\
& g_3(\x)=1-\frac{140.45 x_1}{x_2^2 x_3} \le 0, \\
& g_4(\x)=\frac{x_1+x_2}{1.5} -1 \le 0.
\end{align}
The simple lower and upper bounds are
\be Lb=[0.05, \;\; 0.25, \;\; 2], \quad Ub=[1.00, \;\; 1.30, \;\; 15]. \ee
The best solution found so far is~\citep{Cag2008,Yang2013CSFA}
\be f_{\min}=0.01266522, \quad \x_*=[0.05169, \;\; 0.35673, \;\; 11.2885]. \ee

\item For the three-bar truss system design with two cross-section areas
$x_1=A_1$ and $x_2=A_2$, the objective is to minimize
\be \min f(\x) = 100 (2 \sqrt{2} x_1 + x_2), \ee
subject to
\begin{align}
g_1(\x) & =\frac{(\sqrt{2} x_1 + x_2) P }{\sqrt{2} x_1^2 + 2 x_1 x_2} - \sigma \le 0, \\
g_2(\x) & =\frac{x_2 P}{\sqrt{2} x_1^2 +2 x_1 x_2} -\sigma \le 0, \\
g_3(\x) & =\frac{P}{x_1 +\sqrt{2} x_2}-\sigma \le 0.
\end{align}
where $\sigma=2$ kN/cm$^2$ is the stress limit and $P=2$ kN is the load.
In addition, $x_1, x_2 \in [0, 1]$.

The best solution so far in the literature is~\citep{Bekdas2018}
\be f_{\min}=263.8958, \quad \x_*=(0.78853, 0.40866). \ee

\item Beam design. For the beam design to support
a vertical load at the free end of the beam, the objective is to minimize
\be \min f(\x)=0.0624 (x_1+x_2+x_3+x_4+x_5), \ee
subject to
\be g(\x)=\frac{61}{x_1^3} + \frac{37}{x_2^3} + \frac{19}{x_3^3}
+\frac{7}{x_4^3} + \frac{1}{x_5^3}-1 \le 0, \ee
with the simple bounds $0.01 \le x_i \le 100$.
The best solution found so far in the literature is~\citep{Bekdas2018}
\be f_{\min}=1.33997, \quad \x_*=(6.0202, 5.3082, 4.5042, 3.4856, 2.1557). \ee

\item Pressure vessel design. The main objective is to minimize
\be \min f(\x)=0.6224 x_1 x_3 x_4 +1.7781 x_2 x_3^2 +3.1661 x_1^2 x_4+19.84 x_1^2 x_3,  \ee
subject to
\begin{align}
& g_1(\x)=-x_1+0.0193 x_3 \le 0, \\
& g_2(\x)=-x_2+0.00954 x_3 \le 0, \\
& g_3(\x)=-\pi x_3^2 x_4 -\frac{4 \pi}{3} x_3^3+1 296 000 \le 0, \\
& g_3(\x)=x_4-240 \le 0.
\end{align}

The first two design variables must be the integer multiples of
the basic thickness $h=0.0625$ inches~\citep{Cag2008}. Therefore,
the lower and upper bounds are
\be Lb=[h, \; h, \; 10, \; 10], \ee
and \be  Ub=[99h, \; 99h, \; 200, \; 200]. \ee

Its true global optimal solution $f_{\min}=6059.714335$ occurs at
\be x_1=0.8125, \;\; x_2=0.4375, \; \; x_3=40.098446, \; \;
x_4=176.636596. \ee

\item Parameter estimation of an ODE.
For a vibration problem with a unit step input, we have its mathematical equation as an ordinary differential equation~\citep{Yang2023ten}
\be \frac{\ddot{y}}{\omega^2} + 2 \zeta \frac{\dot{y}}{\omega} + y= u(t), \label{ODE-eq-100} \ee
where $\omega$ and $\zeta$ are the two parameters to be estimated.
Here, the unit step function is given
\be u(t)=\brak{0 & \quad \textrm{if } \;\; t<0, \\ 1 & \quad \textrm{if } \;\; t \ge 0. } \ee
The initial conditions are $y(0)=y'(0)=0$. For a given system, we have observed its actual response.
The relevant measurements are given Table~\ref{Table-data-100}.
\begin{table}
\begin{center}
\caption{Measured data for a vibration problem. \label{Table-data-100}}
\begin{tabular}{|r|rrrrr|}
\hline
Time $t_i$ & 1 & 2 & 3 & 4 & 5\\
$y(t_i)$ & 1.0706 & 1.3372 & 0.8277 & 0.9507 & 1.0848 \\ \hline
Time $t_i$ & 6 & 7 & 8 & 9 & 10 \\
$y(t_i)$  & 0.9814 & 0.9769 & 1.0169 & 1.0012 & 0.9933 \\ \hline
\end{tabular}
\end{center}
\end{table}

In order to estimate the two unknown parameter values $\omega$ and $\zeta$, we can define the objective function as
\be f(\x)=\sum_{i=0}^{10} [y(t_i)-y_s(t_i)]^2, \ee
where $y(t_i)$ for $i=0,1,...,10$ are the observed values and
$y_s(t_i)$ are the values obtained by solving the differential equation~(\ref{ODE-eq-100}),
given a guessed set of values $\zeta$ and $\omega$.
Here, we have used $\x=(\zeta, \; \omega)$.

The true values are $\zeta=1/4$ and $\omega=2$. The aim of this benchmark is to solve the differential equation iteratively so as to find the best parameter values that minimize the objective or best-fit errors.

\end{enumerate}

\subsection{Parameter Settings}

In our simulations, the parameter settings are:
population size $n=10$ with parameter values of
$a=1$, $b=0.7$, $c=1$, $p=0.7$, $q=r=1$ and $\Theta=\theta^t$ with $\theta=0.97$. The maximum number of iterations is set to
$t_{\max}=1000$.

For the benchmark functions, $D=5$ is used for $f_1$, $f_2$, $f_3$,
$f_4$, and $f_{10}$. For $f_9$, $D=4$ is used so as to give $f_{\min}$ as
a nice integer. For all other problems, their dimensionality has been
given earlier in this section.

\section{Numerical Experiments and Results}

After implementing the algorithm using Matlab, we have carried out
various numerical experiments. This section summarizes the results.

\subsection{Results for Function Benchmarks}

For the first ten functions, the algorithm has been run 20 times
so that the best $f_{\min}$, mean values and other statistics can be
calculated. The results have been summarized in Table~\ref{Table-fs-100}.
As we can see, the algorithm can find all the true optimal solutions
even with a small population size $n=10$. This shows that the unified GEM algorithm is very efficient for solving function optimization problems.

\begin{table}[ht]
\begin{center}
\caption{Numerical experiments for ten test functions.  \label{Table-fs-100}}
\begin{tabular}{|l|r|r|r|r|}
\hline
      & Best ($f_{\min}$) & Worst  & Mean ($f_{\min}$) & True $f_{\min}$ \\
\hline
Fun 1 & $+1.4073e-28$  &  $+1.3985e-27$  &  $+5.2722e-28$ & 0 \\
Fun 2 & $+1.0192e-26$  &  $+4.3595e-05$  &  $+4.3595e-06$ & 0 \\
Fun 3 & $+2.6645e-15$  &  $+2.3981e-14$  &  $+9.7700e-15$ & 0 \\
Fun 4 & $+9.8139e-27$  &  $+6.6667e-01$  &  $+2.0000e-01$ & 0 \\
Fun 5 & $-3.4560e+03$  &  $-3.4560e+03$  &  $-3.4560e+03$ & -3456 \\
Fun 6 & $+3.1554e-30$  &  $+1.8302e-28$  &  $+7.9359e-29$ & 0 \\
Fun 7 & $-1.9209e+01$  &  $-1.6268e+01$  &  $-1.8620e+01$ & -19.2085 \\
Fun 8 & $+1.9586e-29$  &  $+7.6207e-01$  &  $+7.6207e-02$ & 0 \\
Fun 9 & $-1.6000e+01$  &  $-1.6000e+01$  &  $-1.6000e+01$ & -16 \\
Fum 10 & $+0.0000e+00$  &  $+1.9899e+00$  &  $+1.1940e+00$ & 0 \\
\hline
\end{tabular}
\end{center}
\end{table}

\subsection{Five Case Studies}

To test the proposed algorithm further, we have used it to solve five different case studies,  subject to various constraints. The constraints are handled by the standard penalty method with a penalty coefficient $\lam=1000$ to $10^5$. For the pressure vessel problem, $\lam=10^5$ is used, whereas
all other problems use $\lam=1000$.

For example, in the pressure vessel design problem, the four constraints are
incorporated as $P(\x)$ so that the new objective becomes
\be F_p(\x)=f(\x)+\lam P(\x), \ee
where
\be P(x)=\sum_{i=1}^4 \max\{0, g_i(\x)\}, \quad \lam=10^5.  \ee
The pressure vessel design problem is a mixed integer programming problem because the first two variables $x_1$ and $x_2$ can take only integer multiples of the basic
thickness $h=0.0625$ due to manufacturing constraints. The other two
variables $x_3$ and $x_4$ can take continuous real values. For each
case study, the algorithm has been run 10 times. For example, the 10 runs
of the pressure vessel design are summarized in Table~\ref{Table-PV-100}.
As we can see, Run 2 finds a much better solution $f_{\min}=5850.3851$
than the best solution known so far in the literature $6059.7143$.
All the constraints are satisfied, which means that this is a valid new solution.

\begin{table}[ht]
\begin{center}
\caption{Pressure vessel design benchmarks. \label{Table-PV-100}}
\begin{tabular}{|l|r|r|r|r|r|}
\hline
      & $x_1$ & $x_2$ & $x_3$ & $x_4$ & $f_{\min}$ \\
\hline
Run 1 & $ 1.2500$  &  $ 0.6250$  &  $64.7668$  &  $11.9886$  &  $7332.8419$ \\
Run 2 & $ 0.7500$  &  $ 0.3750$  &  $38.8601$  &  $221.3657$  &  {\bf  $5850.3851$} \\
Run 3 & $ 1.1875$  &  $ 0.6250$  &  $61.5285$  &  $26.9310$  &  $7273.5127$ \\
Run 4 & $ 1.0000$  &  $ 0.5000$  &  $51.8135$  &  $84.5785$  &  $6410.0869$ \\
Run 5 & $ 0.8750$  &  $ 0.4375$  &  $45.3367$  &  $140.2544$  &  $6090.5325$ \\
Run 6 & $ 0.8750$  &  $ 0.4375$  &  $45.3368$  &  $140.2538$  &  $6090.5262$ \\
Run 7 & $ 0.7500$  &  $ 0.3750$  &  $38.8601$  &  $221.3655$  &  $5850.3831$ \\
Run 8 & $ 1.0000$  &  $ 0.5625$  &  $51.8135$  &  $84.5785$  &  $6708.4337$ \\
Run 9 & $ 0.8125$  &  $ 0.4375$  &  $42.0984$  &  $176.6366$  &  $6059.7143$ \\
Run  10 & $ 0.7500$  &  $ 0.4375$  &  $38.8601$  &  $221.3655$  &  $6018.2036$ \\
\hline
\end{tabular}
\end{center}
\end{table}

Following the exact same procedure, each case study has been simulated 20 times. The results for the five design case studies are summarized in Table~\ref{Table-Sim-200}. As we can see, all the best known optimal
solutions have been found by the algorithm with a population size $n=10$.

\begin{table}[ht]
\begin{center}
\caption{Results for the five design benchmarks. \label{Table-Sim-200}}
\begin{tabular}{|l|r|r|r|r|}
\hline
      & Best ($f_{\min}$) & Worst & Mean ($f_{\min}$) & Best so far \\
\hline
Spring Design & $1.2665e-02$  &  $1.2873e-02$  &  $1.2729e-02$
& $1.2665e-02$ \\
Truss System & $2.6389e+02$  &  $2.6390e+02$  &  $2.6390e+02$
& $2.6389e+02$ \\
Beam Design & $1.3400e+00$  &  $5.1657e+00$  &  $2.3158e+00$
& $1.3400e+00$ \\
Pressure Vessel & $5.8504e+03$  &  $7.3328e+03$  &  $6.3685e+03$
& $6.0597e+03$ \\
ODE Parameters & $6.9479e-09$  &  $1.5850e-02$  &  $3.1700e-03$
& $6.8900e-09$ \\
\hline
\end{tabular}
\end{center}
\end{table}

The above simulations and results have shown that the unified GEM algorithm performs well for all the 15 test benchmarks, and in some cases it can achieve even better results. This indicates that this unified algorithm
is effective and can potentially be applied to solve a wide range of optimization problems. This will form part of our further research.

\section{Conclusion and Discussion}

In this paper, we have proposed a unified algorithm, called generalized
evolutionary metaheuristic (GEM), which represents more than 20 different algorithms in the current literature. We have validated the proposed GEM with 15 different test benchmarks and optimal solutions have been obtained
with a small population size and fixed parameters.

From the parameter tuning perspective, it can be expected that if the parameters can be tuned systematically, it may be possible to enhance the algorithm's performance further. In fact, a systematical parameter study
and parameter tuning is needed for this new algorithm,
which will be carried out in our future work.

In addition, apart from more than 20 algorithms as special cases of
this unified algorithm when setting different parameter values, it can be
expected that the vast majority of the 540 algorithms can also be rightly represented in this unified algorithm if certain parameter are allowed to vary in certain ways and some minor differences in some algorithms can be ignored. Obviously, for some algorithms such as the genetic algorithm, the algorithm is mainly described as an algorithmic procedure without explicit
mathematical formulas. Such type of algorithm may not be easily categorized  into the generalized framework. However, the procedure and algorithmic flow may in essence be quite similar to some of the main framework. In addition, a systematic comparison of this general framework can be carried out with various component algorithms. This can be a useful topic for further research.

\bibliographystyle{apalike}

\end{document}